\documentclass[journal,twoside,web]{ieeecolor}
\usepackage{generic}
\usepackage{cite}
\usepackage{amsmath,amssymb,amsfonts}
\usepackage{graphicx}
\usepackage{textcomp}
\usepackage[caption=false,font=footnotesize]{subfig}
\usepackage{breqn}
\usepackage[]{algorithm2e}

\SetKwInput{Kw}{Input}

\SetKwInput{KW}{Output}

\usepackage[section]{placeins}
\def\BibTeX{{\rm B\kern-.05em{\sc i\kern-.025em b}\kern-.08em
    T\kern-.1667em\lower.7ex\hbox{E}\kern-.125emX}}
\begin{document}
\title{A Precision Environment-Wide Association Study of Hypertension via Supervised Cadre Models}
\author{Alexander New, \IEEEmembership{Member, IEEE} and Kristin P. Bennett
\thanks{This work is supported by IBM Research AI through the AI Horizons Network and by the Center for Biotechnology \& Interdisciplinary Studies at Rensselaer. }
\thanks{A. N. Author is with the Department of Mathematics and the Institute for Data Exploration and Applications (IDEA), Rensselaer Polytechnic Institute, Troy, NY, 12180 (e-mail: newa@rpi.edu).}
\thanks{K. P. B. Author is with the Department of Mathematics and the Institute for Data Exploration and Applications (IDEA), Rensselaer Polytechnic Institute, Troy, NY, 12180 (e-mail: bennek@rpi.edu).}
}

\maketitle


\begin{abstract}
We consider the problem in precision health of grouping people into subpopulations based on their degree of vulnerability to a risk factor. These subpopulations cannot be discovered with traditional clustering techniques because their quality is evaluated with a supervised metric: the ease of modeling a response variable over observations within them. Instead, we apply the supervised cadre model (SCM), which does use this metric. We extend the SCM formalism so that it may be applied to multivariate regression and binary classification problems. We also develop a way to use conditional entropy to assess the confidence in the process by which a subject is assigned their cadre. Using the SCM, we generalize the environment-wide association study (EWAS) workflow to be able to model heterogeneity in population risk. In our EWAS, we consider more than two hundred environmental exposure factors and find their association with diastolic blood pressure, systolic blood pressure, and hypertension. This requires adapting the SCM to be applicable to data generated by a complex survey design. After correcting for false positives, we found 25 exposure variables that had a significant association with at least one of our response variables. Eight of these were significant for a discovered subpopulation but not for the overall population. Some of these associations have been identified by previous researchers, while others appear to be novel. We examine several discovered subpopulations in detail, and we find that they are interpretable and that they suggest further research questions.
\end{abstract}


\section{Introduction}\label{sec:introduction}

Precision health approaches \cite{precision_healthcare} are increasingly being adopted in health analytics. In these approaches, people with different characteristics are modeled as having different levels of vulnerability and resistance to a harmful condition. For example, \cite{breast} found that, for non-Hispanic Black women, there was a significant association between developing breast cancer and becoming overweight; however, this significant association did not appear in other cohorts of women. Thus, making use of properly chosen subpopulations is important when analyzing complex health problems. When possible, as in the above example, domain knowledge should be used to infer what subpopulations are useful for a given precision health problem. In cases where the existing knowledge is not sufficiently complete to do this, machine learning methods can discover informative subpopulations that lie latent in large heterogenous datasets.

In this work, we develop a novel machine learning method for health risk analysis problems, in which the goal is to identify risk factors that are strongly associated with a chronic health condition in one or more subpopulations. This method is based on the supervised cadre model (SCM), proposed in \cite{SCM}. The SCM discovers subpopulations and assigns each subpopulation a simple model. Sparsity-inducing regularization \cite{lasso} ensures that subpopulations are defined via simple rules -- for example, subjects above a certain threshold based on age and BMI vs. those below it. This means that subpopulations suggested by an SCM are easily validated by a domain expert, granting them a useful form of interpretability \cite{Lipton}.

We can use the SCM for any type of risk analysis, here, we focus on two cases. In the first, the response is the vector of a subject's diastolic and systolic blood pressure readings (DBP and SBP), which we jointly refer to as continuous blood pressure (CBP). Here the risk analysis is based on multivariate regression. In the second, the response is whether or not a subject's blood pressure is high enough to classify them as having hypertension (HYP), so binary classification is used.

The SCM has two primary components. The first is the cadre membership function, which assigns every subject a distribution characterizing that subject's probability of belonging to each cadre. The second component of an SCM is a set of score-prediction models, one for each cadre. These predict the expected response score of that subject, assuming that subjects belongs to a specific cadre. For HYP, the score-prediction models output a scalar risk score. For CBP, the score-prediction functions output a two-dimensional vector containing the predicted systolic and diastolic blood pressures. Thus, the same cadre structure is used for both SBP and DBP predictions, although each has different regression weights.

Importantly, the cadre membership function and score-prediction functions are learned simultaneously. Rather than being chosen to minimize an unsupervised quantity such as within-cluster-sum-of-squares, cadres are selected to maximize the effectiveness of the score-prediction process. Only a small subset of covariates are used for the the cadre-assignment and target-prediction processes -- their functions are sparse with respect to subject characteristics \cite{lasso}. The cadre membership function and each cadre's target-prediction function are allowed to use a different set of covariates.

We use the SCM to carry out an environment-wide association study (EWAS, \cite{PatelDiabetes}). This class of study analyzes the association between one or more response condition(s) -- for us, blood pressure and hypertension -- and many different environmental exposure risk factors, such as trace metals and pesticides. Using the SCM, we generalize existing environmental exposure risk analysis methods. These methods often consider only a small number of possible risk factors, and they are typically restricted to either population-level analysis, or to analysis over a small set of explicitly-chosen subpopulations.

Smaller association studies have been carried out for blood pressure and hypertension \cite{trasande,shiue,bloodcadmium}. Some of these have used our dataset, the National Health and Nutrition Examination Survey (NHANES, \cite{nhanesData}), a publicly accessible, cross-sectional examination of the American population. However, a large-scale hypertension EWAS has not yet been performed.

The SCM discovers twenty-five risk factors that have a significant association with DBP, SBP, or HYP. Eight of these are identified as significant risk factors because we used subpopulation-based modeling. Some of these factors have been identified in previous studies, and others have not. We analyze the subpopulations discovered by the SCM.

\subsection{Related Works}\label{sec:related}

The supervised cadre model learns a soft (i.e., probabilistic) partition of subject-space, and then each element of this partition is assigned an interpretable linear model. This can be viewed as a modification of a simple hierarchical mixture of experts (HME, \cite{jordanHME}). However, the SCM uses a different gating function than the HME. Its parameterization lets the elements of the partition be interpreted as subpopulations centered around a mean-subject.

The SCM is comparable to semi-supervised clustering \cite{semisupervised}.
In semi-supervised clustering, the training set is divided into a set of labeled and unlabeled observations, and supervised and unsupervised metrics are combined to learn a single model. Given a dataset corresponding to a $L$-label classification problem, the $M$-cadre SCM learns a model that assigns to an observation with class $\ell$ a joint class-cadre label from the set $\{(1,\ell),\hdots,(L,\ell)\}$. For interpretability, the SCM uses feature selection in its cadre-assignment procedure. Unsupervised methods have also used feature selection. Examples include the sparse $k$-means \cite{sparse_cluster} and weighted fuzzy $c$-means \cite{fuzzy} methods. These methods solve problems without response variables, and sparse $k$-means focuses on those in the $p\gg n$ regime. Our interest is supervised learning problems, as our goal is the discovery of subpopulations that provide useful information about the variation of a response variable.

When performing risk analysis with the SCM, we want discovered subpopulations to be easily validated by a health expert. This requires model interpretability. Recent works such as \cite{Lipton,Holzinger2018,DoshiVelez} have proposed different ways to characterize the degree to which a model is interpretable, such as simulatability, decomposability, and amicability to ante-hoc interpretation. The SCM is simulatable because a human can easily replicate a model's prediction. It is decomposable because all of its parameters have intuitive purposes. It also admits ante-hoc interpretations because variable distributions can be grouped by subpopulation using model parameters for visualization.


\section{Methods}\label{sec:methods}

\subsection{Supervised Cadre Models}\label{sec:scm}

In this section, we describe the mathematical formalism behind the learning and prediction processes for a supervised cadre model. Reference \cite{SCM} only used the SCM for scalar regression problems. In this work, we extend the range of problems the SCM can be applied to. Let a subject be represented by $x \in \mathbb{R}^P$ and let the response be $y \in \mathcal{Y}\subseteq \mathbb{R}^{P_Y}$. For hypertension, $P_Y=1$ and $\mathcal{Y} = \{-1, +1\}$; for continuous blood pressure, $P_Y = 2$ and $\mathcal{Y} = \mathbb{R}^{P_Y}$. In an EWAS, the measurements constituting a vector $x$ contain a set of control variable values (e.g. age and ethnicity) and a single environmental risk factor. Control variables and general experiment design are discussed in detail in Section \ref{sec:variable-selection}.

Let $F_P = \{1,\hdots,P\}$ be the full set of covariate indices. We choose index sets $F_C, F_T \subseteq F_P$, with $P_C = |F_C|$ and $P_T = |F_T|$. If $p \in F_C$, then the covariate $x_p$ is used to determine which cadre a subject $x$ belongs to. If $p \in P_T$, then the covariate $x_p$ is used to predict the response.  In our analysis, we set $F_C$ equal to the set of control variable indices. We set $F_T$ equal to the union of $F_C$ and the index corresponding to the single risk factor.

Let $M$ be the number of cadres in the model. We define a score function $f:\mathbb{R}^P \to \mathbb{R}^{P_Y}$ with the form
\begin{equation}
f(x) = g(x_{F_C})^T e(x_{F_T}),
\label{eq:prediction_function}
\end{equation}
where $g(x_{F_C}) = \begin{bmatrix}
g_1(x_{F_C})\,\cdots\,g_M(x_{F_C}) 
\end{bmatrix}\in\mathbb{R}^M$ and $e(x_{F_T}) = \begin{bmatrix} e^1(x_{F_T})\,\cdots\,e^M(x_{F_T})\end{bmatrix} \in R^{P_y \times M}$. Here, $g_m(x_{F_C})$ is the probability that subject $x$ belongs to cadre $m$, and $e^m(x_{F_T}) \in R^{P_Y}$ is the expected response value for $x$ if $x$ were known to be in cadre $m$.

The SCM imposes parametric forms on $g$ and $e$:
\begin{equation*}
g_m(x_{F_C}) = \frac{e^{-\gamma ||x_{F_C} - c^m||_d^2}}{\sum_{m'}e^{-\gamma ||x_{F_C} - c^{m'}||^2_d}}\,\,\,\,\text{and}
\end{equation*}
\begin{equation*}
e^m(x_{F_T}) = \left(W_m\right)^T x_{F_T} + w_0^m.
\end{equation*}
Here: $||z||_d = \left(\sum_p |d_p| (z_p)^2\right)^{1/2}$  is a seminorm; $d$ is a feature-selection parameter used for cadre assignment; each $c^m \in \mathbb{R}^{P_C}$ is the center of the $m$th cadre; each pair $W_m \in \mathbb{R}^{P_T\times P_Y}, w_0^m \in \mathbb{R}^{P_Y}$ characterizes the regression hyperplane for cadre $m$; and $\gamma > 0$ is a hyperparameter that controls the sharpness of the cadre-assignment process. Thus, the cadre membership of $x$ is a multinoulli random variable with probabilities $\{g_1(x_{F_C}),\hdots,g_M(x_{F_C})\}$; this set is the softmax \cite{Murphy} of the set of weighted inverse-distances \\$\{\gamma||x_{F_C}-c^1||_d^{-2},\hdots,\gamma||x_{F_C}-c^M||_d^{-2}\}$. 

If we let $C = \{c^1,\hdots,c^M\}$, $W = \{W_1,\hdots,W_M\}$, and $W_0 = \{w^1_0,\hdots,w^M_0\}$, the SCM is fully specified by the the parameters $C$, $d $, $W$, $W_0$, and the hyperparameter $\gamma>0$. We group a model's parameters as $\Theta = \{C, d, W, W_0\}$. All the parameters have interpretations, ensuring model decomposability \cite{Lipton}. Each $c^m$ is the center of the $m$th cadre. The coefficient $d_p$ indicates how important the $p$th feature is for the cadre-assignment process. Each cadre has one or more regression hyperplanes characterized by $W_m$ and $w_0^m$.

\subsection{Learning a survey-weighted SCM}\label{sec:learning}

The training process for the survey-weighted SCM is similar for the HYP and CBP cases. In both cases, we first specify a probabilistic model $p(y|x)$, and then we use Bayesian point estimation to learn the model. We model CBP with $$p(y | x) \sim \mathcal{N}(f(x), \Sigma)\,\,\,\,\,\,\,\,\,\,\,\,\,\,\,\Sigma =\begin{bmatrix} \sigma_{sbp}^2 & 0 \\ 0 & \sigma_{dbp}^2\end{bmatrix} \in \mathbb{R}^{P_Y \times P_Y}.$$ For HYP, we adopt the probabilistic model 
$$p(y|x) \propto e^{-\max\{0, 1 - y f(x)\}}.$$ Thus, modeling continuous blood pressure requires an additional parameter compared to hypertension: $\Sigma$. Let $\Theta_{full} = \Theta$ if the response is hypertension, and let $\Theta_{full} = \Theta\cup\{\Sigma\}$ is the response is continuous blood pressure. Let $X=\{x^n\}_{n=1}^N$ be the set of training data, with associated response values $Y=\{y_n\}_{n=1}^N$. Then the optimal parameters $\Theta_{full}$ are a solution to the log-posterior maximization problem
$$\max_{\Theta_{full}} \log p(\Theta_{full} | X, Y) = \log p(Y | X, \Theta_{full}) + \log p(\Theta_{full}).$$

We factor the prior as $p(\Theta_{full}) = p(d|\Sigma)p(W|\Sigma)p(\Sigma)$ when the response is blood pressure, and as  $p(\Theta_{full}) = p(d)p(W)$ for hypertension.
In both cases, we assign $W$ and $d$ elastic-net \cite{ElasticNet} priors to encourage sparse but  stable models. The covariance parameters are given uninformative priors \cite{gelman}: $p(\sigma^2) \propto 1/\sigma^2$ for both $\sigma_{sbp}$ and $\sigma_{dbp}$. 

The SCM learning problem requires the specification of the following hyperparameters: the number of cadres $M$, the cadre-sharpness $\gamma$, the elastic-net mixing weights $\alpha_d$ and $\alpha_W$, and the regularization strengths $\lambda_d$ and $\lambda_W$. Once these are specified, we learn the model via stochastic gradient descent (SGD) in Tensorflow \cite{tensorflow}; the specific solver is Adam \cite{Adam}, which uses adaptive stepsizes for faster convergence. Thus, the final optimization problem for both problems is $$\hat{\Theta}_{full} = \arg\max_{\Theta_{full}} \mathcal{L}(\Theta_{full}),$$
which may be expanded as
\begin{equation}
\begin{split}\label{eqn:hyploss}
\mathcal{L}_{HYP}(\Theta_{full}) = -\sum_{n=1}^N s_n\max\{0, 1 - y_n f_{HYP}(x^n)\} \\ - \,\,\,\frac{\lambda_d}{2}(\alpha_d||d_{HYP}||_1 + (1-\alpha_d)||d_{HYP}||_2^2) \\ - \,\,\,\frac{\lambda_W}{2}(\alpha_W ||W_{HYP}||_1 + (1-\alpha_W)||W_{HYP}||_2^2)
\end{split}
\end{equation}
\begin{equation}
\begin{split}\label{eqn:cbploss}
\mathcal{L}_{CBP}(\Theta_{full}) = -\sum_{n=1}^N \frac{s_n}{2}||y_n -f_{CBP}(x^n)||_{\Sigma^{-1}} \\ - \,\,\,\frac{\lambda_d}{2|\Sigma|}(\alpha_d||d_{HYP}||_1 + (1-\alpha_d)||d_{HYP}||_2^2) \\ - \,\,\,\frac{\lambda_W}{2|\Sigma|}(\alpha_W ||W_{HYP}||_1 + (1-\alpha_W)||W_{HYP}||_2^2) \\ - \,\,\,(1 + N)\log |\Sigma|,
\end{split}
\end{equation}
where $\mathcal{L}_{HYP}$ is the log-likelihood for the hypertension model, $\mathcal{L}_{CBP}$ is the log-likelihood for the continuous blood pressure model, $f$ is defined in \eqref{eq:prediction_function}, $|\Sigma| = \sigma_{sbp}^2\sigma_{dbp}^2$ is the determinant of the covariance, and $s_n$ is the survey weight for the $n$th subject. Section \ref{sec:survey_math} describes the need for survey weights in greater detail. The difference in priors between $p(d)$ and $p(W)$ for $\mathcal{L}_{HYP}$ and $p(d|\Sigma)$ and $p(W|\Sigma)$ comes from the derivation of the Bayesian Elastic Net\cite{BayesElasticNet} for regression problems. The SCM learning problem is nonconvex and non-differentiable, but \cite{SCM} reports that, when trained using SGD, all discovered local minimizers tend to be of comparable quality. We consistently find high-quality minimizers in practice.  


\subsection{Risk Analysis on Survey Data}\label{sec:survey_math}

	In this paper, we solve risk analysis problems that are formulated as binary classification and multivariate regression tasks. The goal is to use statistical modeling and significance tests to identify covariates that are strongly associated with the response in one or more of the subpopulations discovered with the SCM. We do not expect to learn models with low out-of-sample error, because hypertension is a complex phenomenon that is affected by lifestyle, dietary, environmental, and genetic factors \cite{trasande}. NHANES, however, has no genetic variables, and its lifestyle and dietary variables are based on questionnaires, which makes them more likely to be biased \cite{PatelDiabetes}.

Each wave of NHANES is constructed via a complex survey design (CSD, \cite{survey_analysis}). In the NHANES CSD, the survey population is divided according to a multistage sample design based on counties and households, and certain minority subpopulations, like the elderly, are oversampled relative to their absolute size in the population. A given subject's role in the NHANES CSD is captured by their survey weight, survey stratum, and survey variance unit. More details about complex survey designs can be found in texts such as~\cite{survey_analysis}.

When analyzing data generated by a CSD, it is important to use appropriate modeling techniques. Specifically, observations in a CSD are not sampled $iid$, and the use of modeling techniques that assume $iid$ will produce biased results. Incorporation of survey weights is necessary to attain correct parameter point estimates, and incorporation of strata and variance units is necessary for valid standard error estimates and confidence intervals for these point estimates. We incorporate survey weights into the SCM loss function \eqref{eqn:hyploss} and \eqref{eqn:cbploss} with the $s_n$ terms. Incorporation of survey strata and variance units for valid standard error estimation requires the use of survey-weighted generalized linear models (GLMs), which are implemented in the \texttt{survey} package for R \cite{survey_package}. The process by which we apply survey-weighted models to subpopulations discovered by the SCM is described in Section \ref{sec:study_design}.

\subsection{Applying Conditional Entropy to Assess Cadre Hardness}\label{sec:entropy}

We discuss how the soft partition learned by an SCM is combined with survey-weighted modeling. Given an SCM, we take every subject and assign them to their most likely cadre, ``hardening'' the soft, probabilistic cadre assignments. 

We evaluate the validity of this simplification with a novel application of conditional entropy \cite{Murphy}. In order for the simplification to be valid, any given subject must be very confidently assigned to its most likely cadre. That is, for any subject $x$, we desire $g_m(x) \approx 0$ or $g_m(x) \approx 1$. Conditional entropy describes the extent to which this condition holds; it measures how ``hard'' a probabilistic partition is.

Let $X_m$ be the set of subjects $x$ such that $g_m(x) \geq g_{m'}(x)$ for $m' \in \{1,\hdots,M\}$. Let $C \in \{1,\hdots,M\}$ be the random variable with conditional probability mass function $p(C = m|x) = g_m(x) = p(\{x \in X_m\})$, for any subject $x$. That is, $C$ is the random variable of conditional cadre assignments. Then consider the conditional entropy 
\begin{equation}
\label{eq:entropy}
H(C|x(m)) = -\sum_{m'}p(m'|x(m))\log_2 p(m'|x(m)),
\end{equation}
where $x(m)$ indicates the event that observation $x$ belongs to cadre $m$, that is, $H(C|x(m)) = H(C | \{x\in X_m\})$. This entropy quantifies how confident the assignment of subjects into cadre $X_m$ tends to be. If $H(C|x(m))$ is close to 0, then cadre assignments are very confident. As $H(C|x(m))$ approaches its maximum value of $\log_2 M$, cadre assignments become less confident, and our deterministic partition approximation becomes less valid.

\section{Data and Variables}\label{sec:data}

We used the National Health and Nutrition Examination Survey (NHANES), a publicly accessible, cross-sectional examination of the American population. NHANES is administered every two years. We used data from 1999 through 2013. 

NHANES divides its variables into components. W used demographic variables (for subject age, gender, ethnicity, and socioeconomic status), examination variables (for subject BMI), and laboratory variables (for environmental exposure variables). We draw potential risk factors from sixteen classes of environmental exposure variables, such as arsenics and polyaromatic hydrocarbons. The sixteen classes are subdivided further into 38 categories for reasons described in Section \ref{sec:variable-selection}. The full list of categories is in the Supplement A.

\subsection{Choosing Variables}\label{sec:variable-selection}

We defined our response variables as follows. An NHANES participant has their blood pressure taken at least three times. As in \cite{trasande}, we averaged each participant's blood pressure readings and used mean systolic and diastolic blood pressure readings as a vector response. As in \cite{shiue}, We  defined a binary response variable for  hypertension (HYP)  by saying that a subject has hypertension if their average systolic BP is at least 140 mmHg and their average diastolic BP is at least 90 mmHg.

When estimating the association of a potential risk factor with a response variable, it is common to control for known confounders, mediating their effect on the response variable\cite{survey_analysis}. For blood pressure, we followed \cite{shiue} and controlled for age, sex, ethnicity, and body mass index. For variables measured in subjects urine, we also controlled for urinary creatinine  \cite{shiue,trasande}. These form the study's control variables.

We extract 218 environmental exposure potential risk factors for analysis, grouped into 38 categories. We need to define these categories because, in NHANES, a given environmental exposure variable will not, in general, be measured on an entire wave's participants, nor will it be measured in every wave. This gives the full NHANES dataset a block-sparse structure. To account for this structure, we divide the classes of risk factors into categories that have all been measured in the same subjects during the same waves. Supplement A gives the number of subjects and risk factors in each category, as well as the waves in which it is present.


\section{Significance Results and Subpopulation Exploration}\label{sec:signif_results}

\subsection{Study Design}\label{sec:study_design}

We summarize our study design with Algorithms \ref{alg:cadre} and \ref{alg:glm}. The former describes the process by which our supervised cadre models were trained. We used grid search to choose the optimal $\lambda_d$ and $\lambda_W$; the metric for model-goodness was the Bayesian Information Criterion (BIC). When choosing the best SCM, we restricted our attention to models that had cadre-assignment conditional entropies \eqref{eq:entropy} of no more than 0.2 for all $m$. This ensures that the deterministic approximation of the soft partition is not unrepresentative. To allow the conditional entropies to be small, we set the cadre-sharpness hyperparameter $\gamma$ to 75. Because we want sparse cadre structures, we favor $\ell^1$ regularization and set the elastic-net mixing hyperparameters $\alpha_d$ and $\alpha_W$ to 0.9. We learn SCMs with $M = 1, 2, 3$ different cadres: forays into larger values of $M$ yielded models with prohibitively high BIC scores, suggesting a saturation of the dataset with respect to model complexity. Note that the model selection and learning is done independently for every risk factor.

\begin{algorithm}
\caption{SCM learning and selection}\label{alg:cadre}
\Kw{\texttt{Response-variable}, \texttt{control-variables}, set of risk factor \texttt{categories}, number of cadres $M$, grid of values for $(\lambda_d, \lambda_W)$, cadre-assignment sharpness $\gamma$, regularization strengths $\lambda_d,\lambda_W$, survey weights $\{s_n\}$}
\KW{Set of optimal models}
\For{every \texttt{category} of risk factors in \texttt{categories}}{
Log transform environmental exposure variables\\
Mean-center and standardize data, including response if modeling CBP\\
\For{every \texttt{risk-factor} in \texttt{category}}{
\For{every $\lambda_d,\lambda_W$}{
Learn an SCM to predict \texttt{response-variable} using \texttt{risk-factor} and all \texttt{control-variables} with $M,\lambda_d,\lambda_W,\alpha_d,\alpha_W,\gamma$ as hyperparameters\\
Calculate $H(C|\{x \in X_m\})$ for $m = 1,\hdots,M$
}
Store model with minimal $BIC$ over all $\lambda_d,\lambda_W$ for \texttt{risk-factor} and $M$, subject to entropy constraints
}
}
\end{algorithm}

Once subpopulations have been discovered with Algorithm \ref{alg:cadre}, Algorithm \ref{alg:glm} describes how survey-weighted GLMs estimate the association between each risk factor and each subpopulation. The strength of this association is captured by the risk factor's regression coefficient log-odds ratio. We use both SCMs and GLMs because survey-weighted GLMs are required to obtain statistically valid parameter standard errors. For HYP, the GLM is logistic regression. For SBP and DBP, the GLM is linear regression. Separate GLMs are learned for the two types of continuous blood pressure because the multivariate regression problem becomes decomposable. Given a learned GLM, we care about three quantities: the $p$-value, regression coefficient, and the regression coefficient's standard error associated with that GLM's environmental exposure variable. We restrict our attention to environmental exposure variables with positive regression coefficients.

Our method generates a very large number of GLMs, and each GLM performs a hypothesis test to assess the significance of that GLM's risk factor. Because so many hypothesis tests are being performed, we perform Benjamini-Hochberg false discovery rate (FDR) correction \cite{Benjamini} on the $p$-values before assessing significance at a threshold of 0.02. All $p$-values reported in the subsequent sections are post-adjustment.

\begin{algorithm}
\caption{Subpopulation-based survey-weighted generalized linear modeling}\label{alg:glm}
\Kw{\texttt{Response-variable}, \texttt{control-variables}, set of risk factor \texttt{categories}, number of cadres $M$, set of optimal cadre models from Algorithm \ref{alg:cadre}, survey variables}
\KW{Set of survey-weighted generalized linear models for each cadre}
\For{every \texttt{category} of risk factors in \texttt{categories}}{
\For{every \texttt{risk-factor} in \texttt{category}}{
\For{$m = 1, \hdots, M$}{
Select all subjects $X_m$ belonging to cadre $m$ from \texttt{risk factor}'s optimal SCM\\
Learn and store survey-weighted GLM on $X_m$ using \texttt{risk-factor} and all \texttt{control-variables}
}
}
}
\end{algorithm}

\subsection{Summary of Results}\label{sec:results_summary}

\begin{figure*}
\begin{center}
\includegraphics[width=0.7\linewidth]{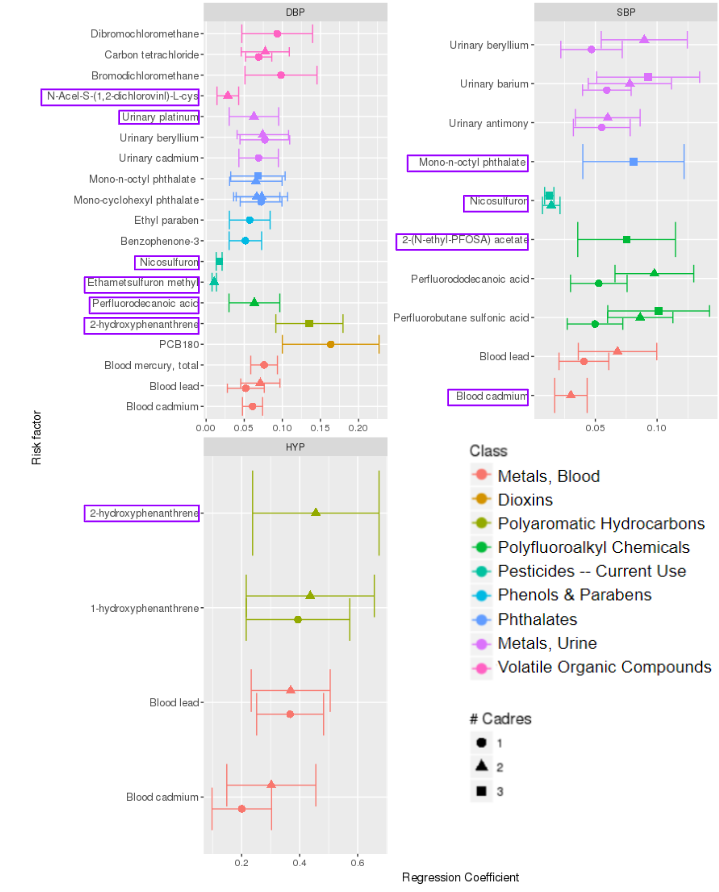}
\caption{\small Distribution of regression coefficients for hypertension corresponding to risk factors that are significant at the (adjusted) $\alpha=0.02$ level, shown with 95$\%$ confidence intervals. Boxed variables indicate significant associations that are only found because subpopulation-level modeling was utilized. The results associated with blood cadmium and 2-hydroxyphenanthrene are explored in greater depth in Sections \ref{sec:cadmium} and \ref{sec:hyd}.}\label{full-logodds}
\end{center}
\end{figure*}

First we summarize the study's results. In Section \ref{sec:subpopulations}, we will examine specific subpopulations. Of the 218 risk factors we considered, 25 had a significant positive association with at least one response variable at an $\alpha = 0.02$ significance threshold. Eleven significant positive associations and eight unique risk factors would not have been identified had we only modeled risk on a population-level. We take the regression coefficients and log-odds ratios corresponding to significant risk factors and plot them in Fig. \ref{full-logodds}. The names of the significant risk factors are listed in Table \ref{table:hits}.

Some risk factors, such as the blood cadmium, are significant for more than one of our three response variables. Blood lead (LBXBPB) and blood cadmium (LBXBCD) are significant for all three response variables. In addition, 2-hydroxyphenanthrene (URXP07) is significant for HYP and DBP, and nicosulfuron (URXNOS), mono-n-octyl phthalate (URXMOP), and urinary beryllium (URXUBE) are significant for DBP and SBP. Some of these associations have been discovered in previous studies. For example, \cite{bloodcadmium} found a positive association between blood cadmium and systolic and diastolic blood pressure for women, \cite{bloodlead} found a positive association between blood lead and SBP, DBP, and HYP in women, and \cite{pressure_phenanthrene} found a positive association between high blood pressure and 2-hydroxyphenanthrene (URXP07). In addition, \cite{PAH_PAD} found positive associations between both 1-hydroxyphenanthrene and 2-hydroxyphenanthrene and peripheral arterial disease. Thus, the SCM is capable of recovering the findings of multiple previous studies in a single analysis; it can also suggest new possible risk factors.

\begin{table}
\resizebox{0.5\textwidth}{!}{
\begin{tabular}{llll}
Code    & Category                   & Variable Name                     &  \\\hline
LBXBCD  & Metals, Blood              & Cadmium                           &  \\
LBXBPB  & Metals, Blood              & Lead                              &  \\
URXUBA  & Metals, Urine              & Barium                            &  \\
URXUSB  & Metals, Urine              & Antimony                          &  \\
URXUPT  & Metals, Urine              & Platinum                          &  \\
URXUBE  & Metals, Urine              & Beryllium                         &  \\
URXEMM  & Pesticides – Current Use   & Ethametsulfuron methyl            &  \\
URXNOS  & Pesticides – Current Use   & Nicosulfuron                      &  \\
URXMCP  & Phthalates                 & Mono-cyclohexyl phthalate         &  \\
URXMOP  & Phthalates                 & Mono-n-octyl phthalate            &  \\
URXP07  & Polyaromatic hydrocarbons  & 2-hydroxyphenanthrene             &  \\
URXP06  & Polyaromatic hydrocarbons  & 1-hydroxyphenanthrene             &  \\
LBXPFBS & Polyfluoroalkyl chemicals  & Perfluorobutane sulfonic acid     &  \\
LBXPFDO & Polyfluoroalkyl chemicals  & Perfluorododecanoic acid          &  \\
LBXPFDE & Polyfluoroalkyl chemicals  & Perfluorodecanoic acid            &  \\
LBXEPAH & Polyfluoroalkyl chemicals  & 2-(N-ethyl-PFOSA) acetate         &  \\
URX1DC  & Volatile Organic Compounds & N-acel-S-(1,2-dichlorovinl)-L-cys &  \\
LBXVCT  & Volatile Organic Compounds & Carbon Tetrachloride              & 
\end{tabular}
}
\caption{List of significant exposure variables, their NHANES codes, and their category. Every risk factor with at least one significant association within a subpopulation is listed here.}\label{table:hits}
\end{table}

\begin{figure*}
\begin{center}
\includegraphics[width=0.8\linewidth]{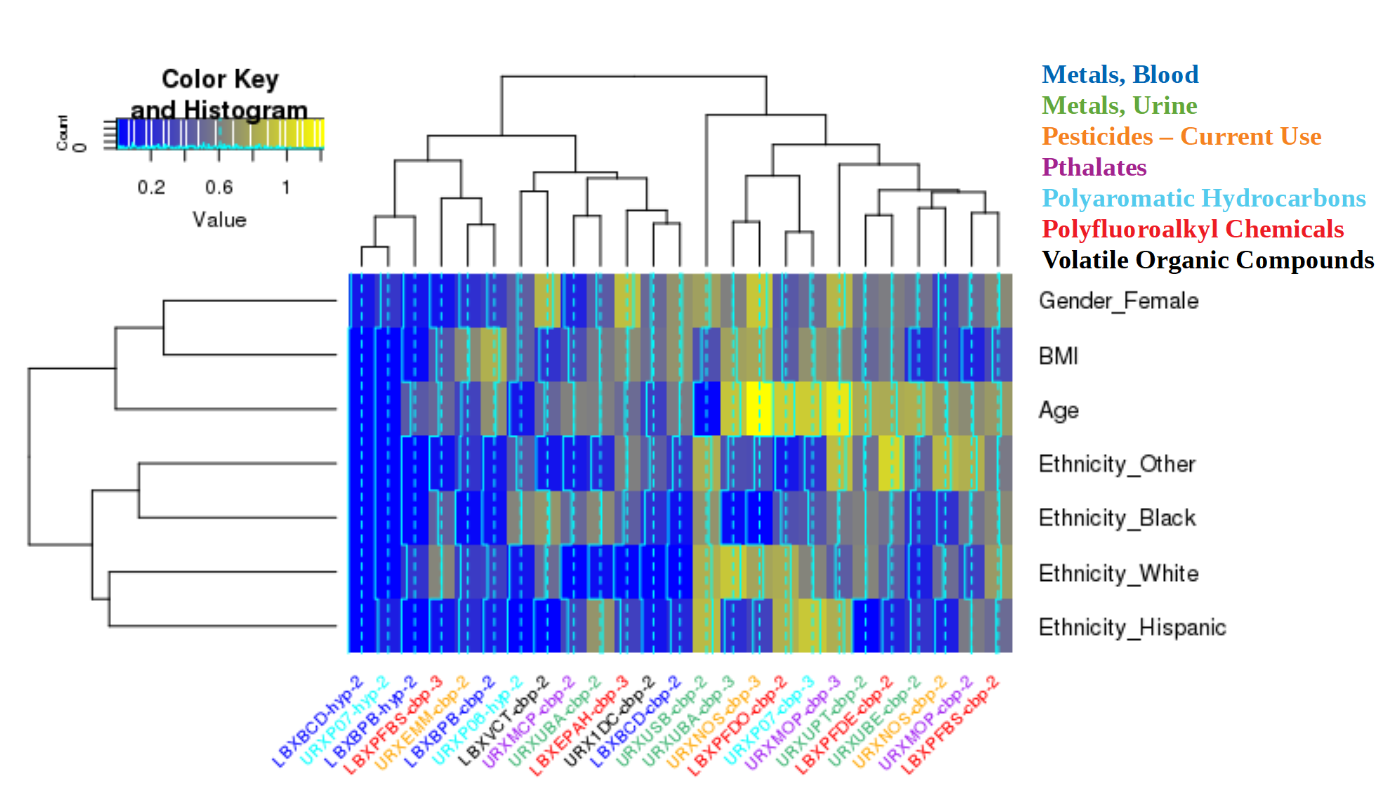}
\caption{\small Cadre-assignment weights $d$ for each SCM that yielded a significant GLM. The column names indicate what model the weight vector $d$ corresponds to: e.g., URXP07-hyp-2 (leftmost column) is the weight vector for the HYP $M=2$ SCM that included URXP07 as a covariate. Rows and columns are sorted by similarity. Solid blue lines help map hues to numeric values in the key, and dotted blue lines show a reference value of 0.5. 
}\label{signif-ds}
\end{center}
\end{figure*}

\subsection{Discovered Subpopulations}\label{sec:subpopulations}

In this section, we explore the subpopulations that the SCM discovers for selected risk factors. We summarize all cadre-assignment weights for SCMs that yield significant associations in Fig. \ref{signif-ds}. Age and gender are frequently important for determining cadre membership, and all ethnicity assignment weights have been grouped together in the hierarchical clustering. We found that many discovered subpopulations contain subjects of only a few ethnicities, which complements prior work \cite{RaceHypertension} finding different rates of hypertension in different ethnicities.

Now we examine individual cadre structure in detail. As we discuss regression coefficients and distributional means, recall that all continuous variables, including SBP and DBP, have been standardized and mean-centered. Throughout, $M$ refers to the number of cadres in the model: saying that a risk factor is significant at the $M=2$ level means that, after an SCM using that risk factor with $M=2$ was trained, at least one of the GLMs for a discovered subpopulations had a significant association between the response and the risk factor.

\subsubsection{Blood Cadmium}\label{sec:cadmium}

In this section, we explore the subpopulations the SCM discovers when blood cadmium is included as a risk factor. Cadmium is a heavy metal that people can be exposed to via tobacco smoke, air pollution, and certain paints. Blood cadmium (LBXBCD, category Metals, Blood) is a significant risk factor on the population-level (i.e., with $M=1$) for DBP (regression coefficient = $0.061\pm0.007$, $p<10^{-8}$). It is a significant risk factor for HYP at the $M=1$ and $M=2$ levels and for SBP at an $M=2$ level. For HYP and SBP, cadre modeling extracts subpopulations with stronger associations between blood cadmium and the response than the general population has.

First we consider the hypertension (HYP) model. At a population level, blood cadmium had a significant association with hypertension (log odds ratio = $0.20\pm0.05$, $p=0.019$). Now we examine the $M=2$ model's subpopulations. In the first (Cadre 1), there was not sufficient evidence to conclude that a significant association between blood cadmium and hypertension exists. In the second (Cadre 2), there was a significant association (log odds ratio $= 0.30\pm0.08$, $p=0.018$). Note that this subpopulation has a larger log-odds ratio than the population does as a whole. Thus, the SCM has pulled out the subjects for whom there is an especially strong association between HYP and blood cadmium. 

Investigation reveals that Cadre 2 was composed exclusively of women. This is similar to the findings of \cite{bloodcadmium}, where blood cadmium had a significant positive association with SBP and DBP for women. Note that  \cite{bloodcadmium} chose to analyze women separately, whereas we recovered this informative subpopulation automatically. There was a significant difference in mean blood cadmium between the two cadres as well (difference = 0.1690448, $p<10^{-10}$). This suggests follow-up questions: Why do women tend to have a higher concentration of blood cadmium, and why is a higher concentration of blood cadmium associated with risk for hypertension in women specifically?

Now we consider the SBP models. In the first (Cadre 1), there was a significant association between blood cadmium and systolic blood pressure (regression coefficient = $0.030\pm0.007$, $p<0.0036$). In the second (Cadre 2), there was not sufficient evidence to find a significant association.

\begin{figure*}
\centering
  \subfloat[BMI ($y$) vs. age ($x$)\label{sbp-lbxbcd-bmi-age}]{%
\includegraphics[width=0.4\linewidth]{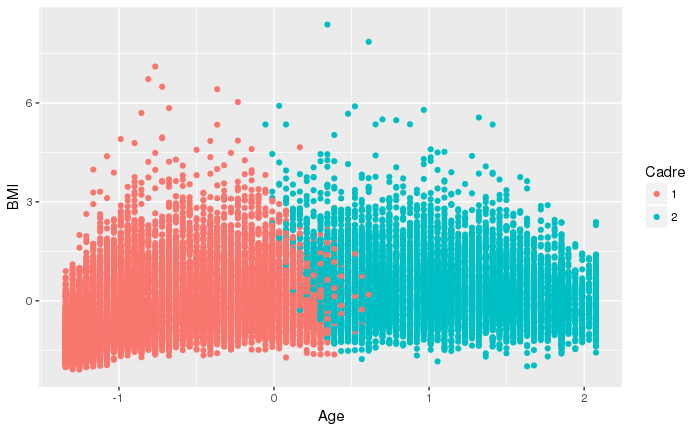}}
    \hfill
\subfloat[Distributions of blood cadmium (left) and SBP (right)\label{sbp-lbxbcd-sbp}]{%
\includegraphics[width=0.4\linewidth]{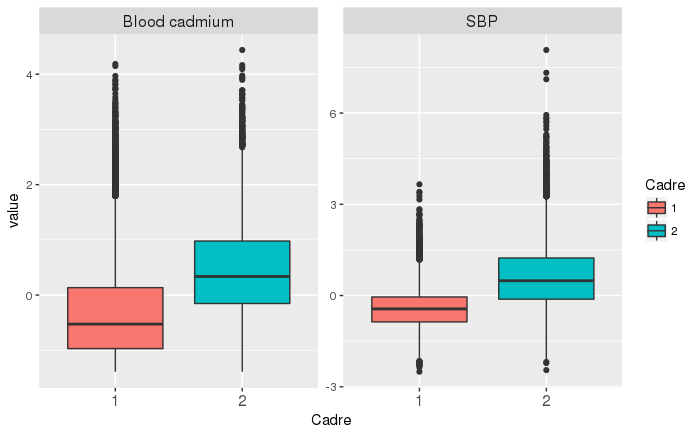}}
\caption{CBP-response blood cadmium subpopulations, colored by subpopulation. Cadre 1 (red) has a significant association between systolic blood pressure and blood cadmium; For Cadre 2 (blue), there was  insufficient evidence to conclude a significant association exists. Cadre 1 is primarily people under the age of 40 who have generally lower systolic blood pressure and blood cadmium values than the members of Cadre 2 do.\label{sbp-lbxbcd}}
\end{figure*}

We visualize the subpopulation structure in Fig. \ref{sbp-lbxbcd}. In Fig. \ref{sbp-lbxbcd-bmi-age}, we see that the subpopulation with a significant association between systolic blood pressure and blood cadmium (Cadre 1) is composed primarily of subjects under the age of 40; for subjects near the age of 40, having a lower BMI makes them more likely to belong to Cadre 1. In Fig. \ref{sbp-lbxbcd-sbp}, we see that subjects in the subpopulation with a significant association between systolic blood pressure and blood cadmium tend to have lower SBP and blood cadmium values than subjects in the other subpopulation do. A interesting question would be to examine why these subjects, who have generally have a lower systolic blood pressure, have a significant association between systolic blood pressure and blood cadmium.

\begin{figure*}
\centering
  \subfloat[Histograms of age, grouped by ethnicity\label{cbp-urxp07-sbp}]{%
\includegraphics[width=0.45\linewidth]{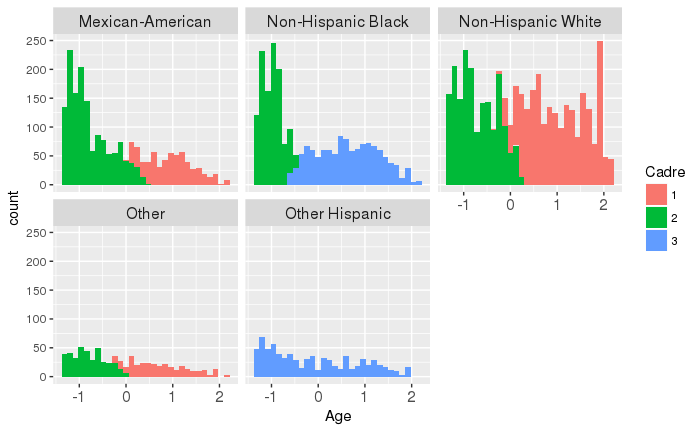}}
    \hfill
\subfloat[Distributions of DBP, SBP, and 2-hydroxyphenanthrene\label{cbp-urxp07-ethnicity-age}]{%
\includegraphics[width=0.45\linewidth]{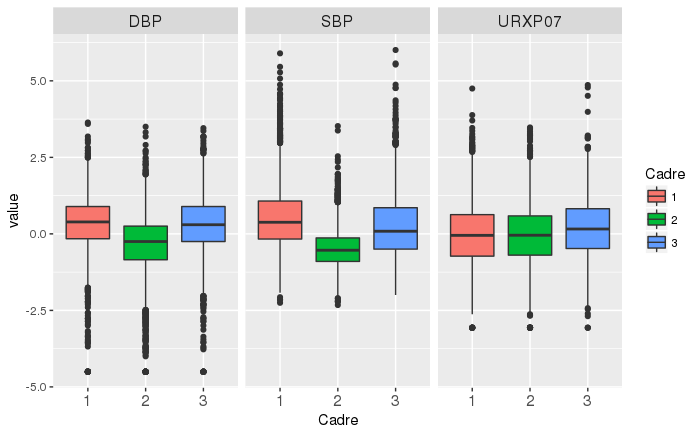}}
\caption{CBP-response 2-hydroxyphenanthrene subpopulations, colored by subpopulation. In cadre 3 (blue), there is a positive significant association between diastolic blood pressure and URXP07; in the other cadres, there was not sufficient evidence to find a significant association. Cadre 3 is Non-Hispanic Black and Other Hispanic people, and its members have higher DBP, SBP, and URXP07  values than cadre 2 (green), which contains younger subjects.\label{cbp-urxp07}}
\end{figure*}

\begin{figure*}
\centering
  \subfloat[Counts of genders and ethnicities\label{hyp-urxp07-gender-ethnicity}]{%
\includegraphics[width=0.45\linewidth]{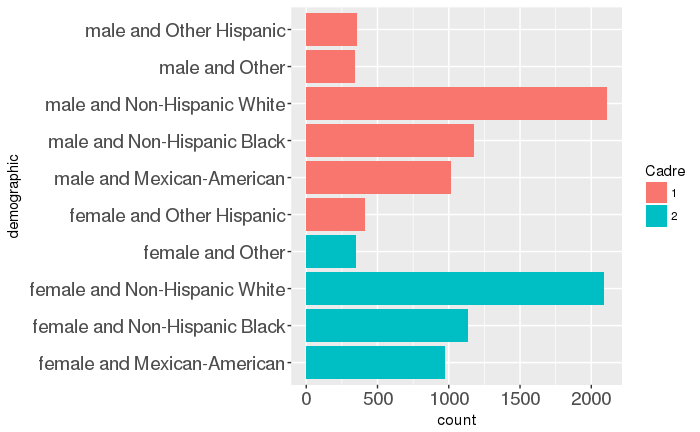}}
    \hfill
\subfloat[Distributions of DBP, SBP, and 2-hydroxyphenanthrene\label{hyp-urxp07-sbp}]{%
\includegraphics[width=0.45\linewidth]{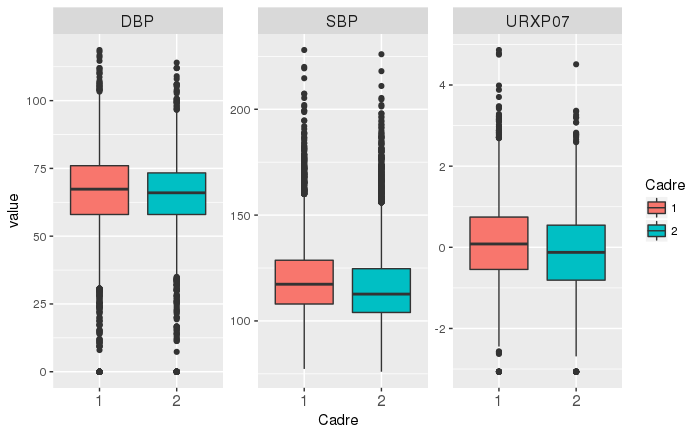}}
\caption{HYP-response 2-hydroxyphenanthrene subpopulations, colored by subpopulation.  In cadre 1 (red), there is a positive significant association between diastolic hypertension and 2-hydroxyphenanthrene; in the other cadre, there was not sufficient evidence to find an association. Cadre 1 is men and Other Hispanic women, and its members have higher DBP, SBP, and 2-hydroxyphenanthrene values than cadre 2 (blue).\label{hyp-urxp07}}
\end{figure*}


\subsubsection{Urinary 2-hydroxyphenanthrene}\label{sec:hyd}

In this section, we explore the subpopulations the SCM discovers when urinary 2-hydroxyphenanthrene (URXP07, category Polyaromatic hydrocarbons) is included as a risk factor. URXP07  is a polyaromatic hydrocarbon -- a class of chemicals that can be produced by the incomplete combustion of organic materials, including coal, tobacco, and food. It is a significant risk factor for DBP with a three-cadre model (regression coefficient = $0.14 \pm 0.02$, $p < 10^{-4}$), and it is a significant risk factor for HYP with a two-cadre model (log-odds ratio = $0.5 \pm 0.1$, $p < 0.009$).


For the DBP-response model, the subpopulation structure is based primarily on age and ethnicity. The subpopulation where a significant association was found (cadre 3) contains Other Hispanic and non-young Non-Hispanic Black subjects, as shown in Fig. \ref{cbp-urxp07-ethnicity-age}. Fig. \ref{cbp-urxp07-sbp} shows that cadre 3 has higher median DBP, SBP, and 2-hydroxyphenanthrene values than cadre 2, which contains younger subjects of all ethnicities. A follow-up analysis could investigate why, despite having similar DBP readings to the subjects in cadre 1, the subjects in cadre 3 have a significant positive association.

For the HYP-response model, the subpopulation structure is based on gender and ethnicity. As show in Fig. \ref{hyp-urxp07-gender-ethnicity}, the subpopulation with a significant association  (cadre 1) contains men of all ethnicities and Other Hispanic women. The subjects in this subpopulation have higher values of DBP, SBP, and 2-hydroxyphenanthrene than the other cadre, as shown in Fig. \ref{hyp-urxp07-sbp}. A follow-up analysis question might be why these men and Other Hispanic women have higher values of DBP, SBP, and 2-hydroxyphenanthrene than women in general do.

\section{Discussion}\label{sec:discussion}

We used a novel supervised cadre model  for a large-scale environmental association study that looked for risk factors associated with high systolic and diastolic blood pressure and hypertension. Our EWAS workflow generalizes the standard EWAS, which is performed only on a population-level. We analyzed more than two hundred risk factors. Twenty-five risk factors had a significant association with at least one response variable; of these, eleven significant associations and eight unique risk factors were discovered due to our use of cadre modeling. Some of our significant associations agreed with other environmental risk factor analyses, while others are novel findings. The SCM learns interpretable subpopulations based on only a small number of covariates, and analysis of these subpopulations suggests further research questions.

Causality is a challenge in association studies such as this one: for an identified association, they cannot determine whether that association is causal or only correlative. However, our use of FDR correction and a low significance threshold means that our significant associations are unlikely to be spurious. Thus, our findings suggest hypotheses to be tested in longitudinal studies and controlled experiments.

Our methods are applicable to types of analysis beyond EWAS: we have demonstrated how the SCM can be applied to general risk analysis problems. We extended the SCM formalism from \cite{SCM} in three ways: (1) the SCM now supports multivariate regression and classification problems, (2) the SCM can be applied to data from complex survey designs, and (3) the uncertainty in the cadre assignment process can be quantified via the estimation of a conditional entropy. Future work could apply this framework to genomics-based precision health problems, such as variant of uncertainty analysis for breast cancer\cite{VUS} or genome-wide association studies\cite{ExampleGWAS}.

\bibliographystyle{IEEEtran}
\bibliography{myBib}

\end{document}